\documentclass{article}

\usepackage[preprint]{corl_2026} %

\usepackage{etoolbox}
\usepackage{enumitem}
\usepackage{xspace}
\usepackage[normalem]{ulem} %
\usepackage{amssymb}
\usepackage{amsmath}
\usepackage{graphicx}
\usepackage{wrapfig}
\usepackage{booktabs}
\usepackage{tabularx}

\usepackage{todonotes}

\def\egowam{{\scshape EgoWAM}\xspace}

\makeatletter
\newcommand\blfootnote[1]{%
  \begingroup
    \renewcommand\thefootnote{}\renewcommand\@makefnmark{}%
    \footnotetext{#1}%
  \endgroup
}
\makeatother

\title{\egowam: World Action Models Beyond Pixels \\ with In-the-Wild Egocentric Human Data}

\author{
  Baoyu Li$^{*}$,
  Xinchen Yin$^{*}$,
  Mengying Lin,
  Yixin Zhang,
  Danfei Xu \\
  Georgia Institute of Technology
}

\begin{document}
\maketitle
\blfootnote{$^{*}$ denotes equal contribution.\quad Correspondence: \texttt{bli678@gatech.edu}.}

\vspace{-10pt}

\begin{abstract}
Egocentric human data offers scalable supervision for robot manipulation. However, behavior cloning entangles transferable content like objects, scenes, and task semantics, with non-transferable factors like human morphology, head motion, and behavioral style. We study whether World Action Models (WAMs) provide a better training signal by requiring policies to predict not only actions, but also how the scene evolves. The central question is what \emph{world representation} best enables human-to-robot transfer. We hypothesize that an effective world target should abstract appearance, capture agent-invariant physical effects, and separate camera motion from environment change. We introduce \egowam, a controlled human-robot co-training framework that fixes the policy backbone, action head, and data mixture while varying only the world prediction target, comparing Pixel, DINO, and 3D motion flow. Across three real-world bimanual tasks, WAM co-training scales more effectively with in-the-wild egocentric human data than behavior cloning. Pixel-based prediction transfers weakly, while DINO and 3D flow yield substantial gains: DINO improves out-of-distribution object and scene generalization by up to 4×, and 3D flow improves in-domain performance by 20–30\%. More details: \href{https://gatech-rl2.github.io/egowam.github.io}{\nolinkurl{gatech-rl2.github.io/egowam.github.io}}.
\end{abstract}

\begin{figure}[!h]
    \centering
    \vspace{-5pt}
    \includegraphics[width=\linewidth]{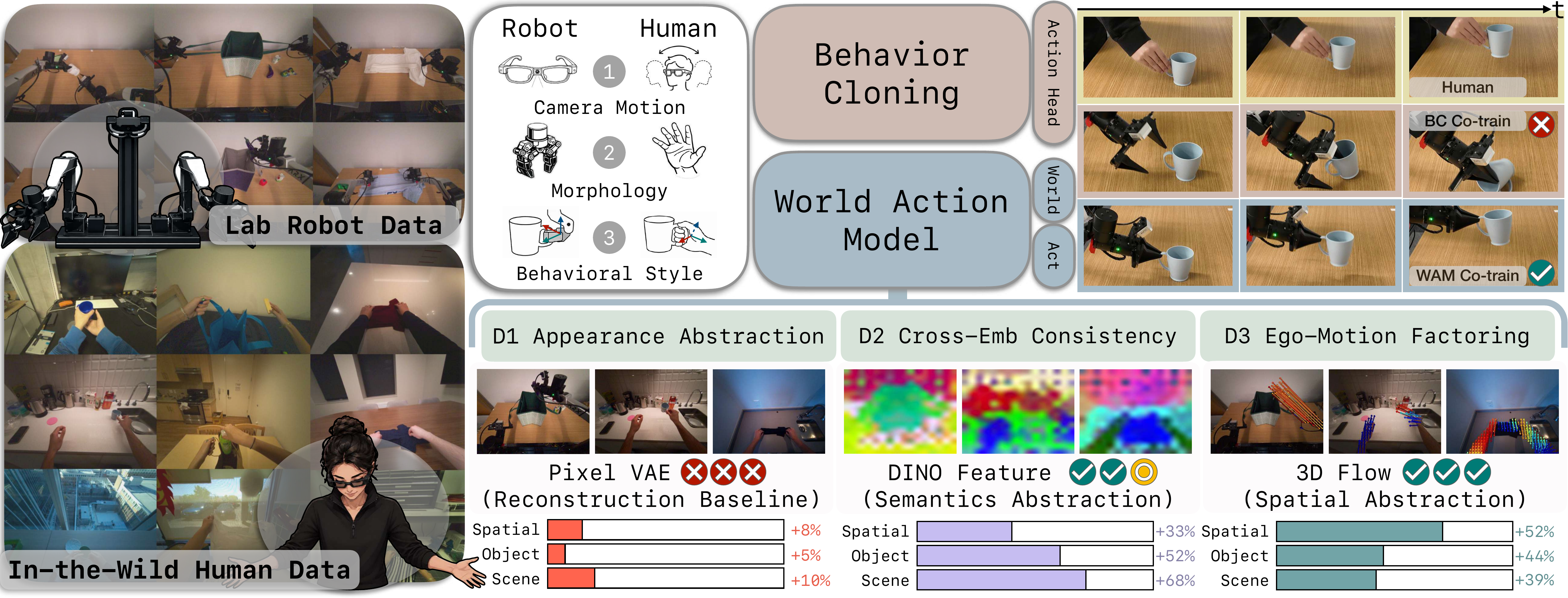}
    \vspace{-15pt}
    \caption{\small
    \textbf{\egowam} co-trains on lab robot data and \textit{in-the-wild egocentric human data}, separated by an embodiment gap of camera motion, morphology, and behavioral style. \textbf{(1)} BC co-training leaks this gap through the \textit{sole} action head as inexecutable motions that harm performance; WAM co-training adds a channel predicting scene evolution, transferring human data through dynamics where actions cannot. 
    \textbf{(2)} \egowam studies \textit{what world representation} best enables transfer, comparing pixel VAE, DINO features, and 3D flow against three desiderata: appearance abstraction, cross-embodiment consistency, and ego-motion factoring.
    }
    \label{fig:teaser}
    \vspace{-10pt}
\end{figure}

\section{Introduction}
\label{sec:introduction}

Egocentric human data offers a promising source for scaling robot manipulation, providing diversity in objects, scenes, and behaviors that is prohibitively expensive to collect on robots~\cite{hoque2026egodexlearningdexterousmanipulation, grauman2024egoexo4dunderstandingskilledhuman, kareer2025emergencehumanrobottransfer}. 
A growing body of work exploits this through behavior-cloning (BC) co-training, retargeting human demonstrations into a robot-compatible action space and jointly training an imitation policy~\cite{kareer2024egomimicscalingimitationlearning, qiu2025humanpolicyhumanoidpolicy}. 
Recent efforts~\cite{zheng2026egoscalescalingdexterousmanipulation, punamiya2026egoverseegocentrichumandataset} show this paradigm can scale, but only with the human data that is carefully aligned to the robot in viewpoint, motion speed, and behavioral style.
Without that bridge, action-level co-training injects human-like motions that the robot cannot execute and \textit{degrades} downstream performance (Fig.~\ref{fig:teaser}). 
This is the \textit{bitter lesson} of action-level co-training: the shared action decoder is the \textit{sole} channel through which human data reaches the policy, forcing it to entangle transferable content (objects, scenes, semantics) with non-transferable execution (morphology, behavioral style), and the embodiment gap in the latter blocks transfer of the former.

World Action Models (WAMs)~\cite{ye2026worldactionmodelszeroshot, kim2026cosmospolicyfinetuningvideo, li2025unifiedvideoactionmodel, zhu2025unifiedworldmodelscoupling} open a second supervision channel: an auxiliary world-model head predicts future states from a shared backbone, grounding action prediction in task-relevant dynamics. Because this channel operates on observations rather than actions, it is largely indifferent to the morphology and behavioral style that vary across embodiments. We hypothesize that task-relevant dynamics transfer across embodiments more readily than actions, so human data can shape the shared backbone through the world-model channel even when its action labels cannot. By decoupling supervision into ``\textit{how the world evolves}'' and ``\textit{how each embodiment acts}'', WAMs induce a more shareable representation and enable more effective scaling from \textit{large-scale in-the-wild human data} than action-level co-training can achieve.

The promise of WAMs for human-robot co-training, however, hinges on a question that has not been systematically studied: \textbf{\textit{what world representation enables effective transfer across embodiments under WAM co-training?}}
Most WAMs predict pixels through a pretrained video VAE~\cite{ye2026worldactionmodelszeroshot, kim2026cosmospolicyfinetuningvideo, li2025unifiedvideoactionmodel, zhu2025unifiedworldmodelscoupling} whose latent is optimized for photometric reconstruction and entangles motion with appearance, which we identify as the dominant failure mode of pixel-level WAM co-training. 
We argue that the world representation should satisfy three desiderata: \textit{appearance abstraction}, \textit{cross-embodiment consistency}, and \textit{ego-motion factoring}.
We then study two alternatives that satisfy them to different degrees: 
DINO features~\cite{oquab2023dinov2, zheng2025diffusiontransformersrepresentationautoencoders}, whose semantic prior abstracts appearance and aligns predictions across embodiments but remains spatially indexed on the image grid, and 3D motion flow~\cite{xiao2025spatialtrackerv23dpointtracking, lu2026track4world}, which satisfies all three by construction through camera-frame-aligned geometric grounding.

We introduce \textbf{\egowam}, a framework for human-robot WAM co-training built upon a Heterogeneous Pretrained Transformer backbone~\cite{wang2024scalingproprioceptivevisuallearningheterogeneous, punamiya2026egoverseegocentrichumandataset}.
\egowam features a shared action head and a \textit{swappable} world-model head trained jointly on human and robot data, isolating the effect of world representation under a matched backbone and data mixture. 
Evaluated on three real-world bimanual tasks spanning spatial, object, and scene generalization, our study makes three contributions:
\begin{itemize}[leftmargin=20pt,topsep=0pt,itemsep=0pt,parsep=0pt]
    \item \textbf{A framework for controlled study of WAM co-training.} \egowam provides a single backbone with a swappable world-model head, enabling apples-to-apples comparison of world representations under identical action supervision and data mixtures.
    \item \textbf{Evidence that WAM co-training unlocks human-data scale.} Across in-domain and OOD evaluations, WAM co-training scales and transfers more consistently from large-scale natural egocentric human data than BC co-training, confirming that future-dynamics supervision is the missing channel where action-only supervision saturates.
    \item \textbf{World representation as the next critical axis.} Both DINO features and 3D motion flow substantially outperform the pixel-VAE baseline, with complementary strengths: DINO's semantic prior yields the strongest object and scene generalization, while 3D flow's geometric grounding yields the strongest spatial generalization and the highest overall scores.
\end{itemize}

\section{Related Work}
\label{sec:related}
\vspace{-5pt}

\noindent \textbf{Robot Learning from Human Data.}
Human video is a scalable data source for robot manipulation~\cite{grauman2024egoexo4dunderstandingskilledhuman, hoque2026egodexlearningdexterousmanipulation, punamiya2026egoverseegocentrichumandataset}. One line of work treats it as a pretraining corpus for action-aware visual representations later transferred to downstream planning or policy~\cite{nair2022r3muniversalvisualrepresentation, assran2025vjepa2selfsupervisedvideo, niu2025arm4r, bai2025wholebodyconditionedegocentricvideo}. Beyond pretraining, prior work uses human data more directly for manipulation along two axes. The first retargets human demonstrations into a shared action space and co-trains an imitation policy~\cite{kareer2024egomimicscalingimitationlearning, qiu2025humanpolicyhumanoidpolicy, yang2025egovlalearningvisionlanguageactionmodels, bi2025hrdthumanmanipulationenhanced, yuan2025motiontranshumanvrdata, li2025vitrascalablevisionlanguageaction} or world model~\cite{goswami2026dexwm, hou2026worldmodelrobotlearning, mendonca2023structuredworldmodelshuman} on both human and robot data; because of domain gaps across embodiments, this hinges on tight alignment in viewpoint, speed, and kinematics~\cite{zheng2026egoscalescalingdexterousmanipulation, li2025h2rhumantorobotdataaugmentation, liu2025immimiccrossdomainimitationhuman, punamiya2025egobridge, cai2025innonscalingegocentricmanipulation, lei2026mechanistic}. The second axis sidesteps the action decoder by extracting object or scene motion from human video and decoding actions from it at inference, especially with 2D point trajectory tracks~\cite{wen2024anypointtrajectorymodelingpolicy, xu2024flowcrossdomainmanipulationinterface, ren2025motiontracksunifiedrepresentation, haldar2025pointpolicykeypoints, lee2025tracegenworldmodeling, collins2025amplifyactionlessmotionpriors} and 3D flow motion fields~\cite{hung2026threepointr3dpointtracks, li2025novaflowzeroshotmanipulationactionable, yin2025objectcentric3dmotionfield, cho2026egoavflowactivevision3dflow,
dharmarajan2025dream2flow}. 
\egowam departs from both: scene motion supervises the shared trunk during \textit{end-to-end} training and is discarded at inference, transferring task-relevant dynamics without injecting inexecutable motions or adding test-time cost.

\noindent \textbf{World Action Models for Robot Learning.}
A world model predicts how the environment evolves given an action~\cite{ha2018worldmodels, hafner2025dreamer4, ai2025review, bar2024navigationworldmodels}; a World Action Model (WAM) couples such a predictor to a policy, grounding action prediction in task-relevant dynamics~\cite{hou2026worldmodelrobotlearning, li2025unifiedvideoactionmodel, zhu2025unifiedworldmodelscoupling}. Most existing WAMs build on a pretrained video model to offer a spatio-temporal representation for action decoding~\cite{ye2026worldactionmodelszeroshot, kim2026cosmospolicyfinetuningvideo, pai2025mimicvideovideoactionmodelsgeneralizable, ma2026dit4ditjointlymodelingvideo, chen2025largevideoplannerenables}, while a parallel line instead treats the video/world model as an interactive simulator for data generation or policy evaluation~\cite{gao2026dreamdojogeneralistrobotworld, wang2026interactiveworldsimulatorrobot, zhang2025real2simeval}. A central but unresolved question for generalizable WAMs is \textit{what world representation} supports scaling from diverse data~\cite{zhou2025dinowmworldmodelspretrained, nilaksh2026reconstructionorsemantics, zhang2024adaptigraph, zhang2024particle, huang2026pointworld, zhang2026molmomotion}. We study this question directly inside the WAM co-training framework, comparing three representations under a single training-as-representation-shaping setup~\cite{yuan2026fastwamworldactionmodels, li2025unifiedvideoactionmodel}: reconstruction-based pixel latents~\cite{wan2025}, semantics-based DINO features~\cite{oquab2023dinov2, zheng2025diffusiontransformersrepresentationautoencoders, singh2026raev2, simeoni2025dinov3}, and camera-stabilized 3D motion flow~\cite{xiao2025spatialtrackerv23dpointtracking, lu2026track4world}.

\section{Human-Robot WAM Co-Training}
\label{sec:wam-cotraining}

\subsection{Aligned Action Channel as a Strong Baseline}
\label{subsec:aligned-action}

We first align the action channel as much as possible so BC co-training is a strong baseline, not a strawman~\cite{punamiya2026egoverseegocentrichumandataset}.
We consider an egocentric human dataset $\mathcal{D}_H = \{(o^H_t, a^H_t)\}_{t=1}^{N_H}$ collected with Project Aria glasses~\cite{engel2023projectarianewtool}, and a bimanual teleoperated robot dataset $\mathcal{D}_R = \{(o^R_t, a^R_t)\}_{t=1}^{N_R}$. Both share an egocentric RGB stream $I^{\text{ego}}$ and the robot additionally has wrist-mounted views $I^{\text{wrist}}$.

We unify cross-embodiment actions into a 14-D end-effector space: per-arm 6-DoF $\mathrm{SE}(3)$ pose plus a 1-D gripper command. Robot actions $a^R_{t:t+k}\!\in\!\mathbb{R}^{k\times d_a}$ are computed from joint angles via forward kinematics and re-expressed in the static ego-camera frame. 
Human hand poses $p^H_{t+i}\!\in\!\mathrm{SE}(3)$ are natively expressed in the moving device frame with transform $T^{\text{device}}_t$; we re-express them in the \emph{instantaneous} device frame at time $t$, $a^H_{t:t+k} = \big[(T^{\text{device}}_t)^{-1} T^{\text{device}}_{t+i}\, p^H_{t+i}\big]_{i=1}^{k}$, factoring out head ego-motion so the same physical motion yields a comparable numerical trajectory across embodiments.

Two residual mismatches remain after coordinate alignment. 
\emph{(i) Speed.} Humans act faster than teleoperated robots, so we use embodiment-specific windows spanning comparable progress---$T_H=1\,\text{s}$, $T_R=1.5\,\text{s}$---both discretized into $k$ steps, yielding semantically aligned trajectories. We use $k$ for resampled chunk length and $T\in\{T_H, T_R\}$ for the original-time horizon, with world-model targets $s_{t+T}$ indexed in original time.
\emph{(ii) Workspace range.} Quantile normalization maps each action dimension's 1st and 99th percentiles to $[-1,1]$, robust to hand-tracking outliers.

Given these aligned inputs, a shared encoder $f_\phi:\mathcal{O}_H\cup\mathcal{O}_R\to\mathcal{Z}$ and a shared action decoder $\pi_\theta(a\mid z)$ are trained end-to-end with the cross-embodiment BC objective
\setlength{\abovedisplayskip}{1pt}
\setlength{\belowdisplayskip}{1pt}
\begin{equation}
\mathcal{L}_{\text{BC-cotrain}}(\phi,\theta) = \sum_{\mathcal{D}\in\{\mathcal{D}_H,\mathcal{D}_R\}} \mathbb{E}_{(o,a)\sim\mathcal{D}}\, \mathcal{L}_{\text{BC}}\!\left(\pi_\theta(a\mid f_\phi(o)),\, a\right),
\label{eq:bc-cotrain}
\end{equation}
realized as the sum of per-embodiment conditional flow-matching losses, $\mathcal{L}_{\text{BC-cotrain}} = \mathcal{L}_{\text{CFM}}^{\text{robot}} + \mathcal{L}_{\text{CFM}}^{\text{human}}$. This is the strongest action-aligned baseline our system supports; the remaining transfer gap motivates the world-model interface introduced next.

\subsection{WAM as a World-Level Transfer Interface}
\label{subsec:wam-interface}

\begin{figure*}[t]
    \centering
    \includegraphics[width=\linewidth]{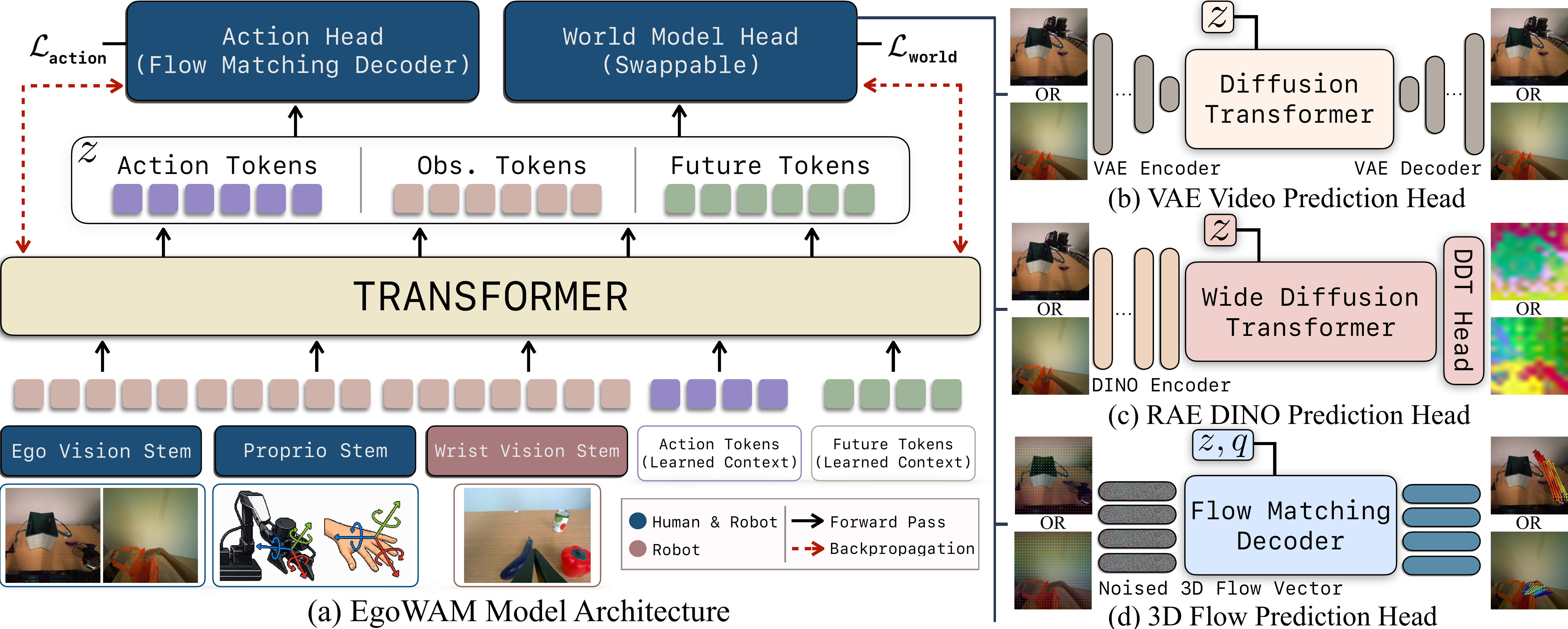}
    \vspace{-16pt}
    \caption{\small
    \textbf{Model Architecture.} \textbf{(a)} \egowam builds on a Heterogeneous Pretrained Transformer~\cite{wang2024scalingproprioceptivevisuallearningheterogeneous, punamiya2026egoverseegocentrichumandataset} with modality-specific stems (ego vision, proprioception, wrist vision), learned action and future tokens, a flow-matching action head, and a swappable world-model head supplying dynamics supervision that carries human data where actions cannot. The head supports three world targets: \textbf{(b)} a VAE head decoding pixel latents, \textbf{(c)} an RAE DINO-feature head, and \textbf{(d)} a 3D-flow head denoising camera-stabilized motion flow.
    }
    \label{fig:method}
    \vspace{-12pt}
\end{figure*}

To open a second supervision channel, we augment the BC policy with an auxiliary future-prediction head, yielding the World Action Model (WAM) family. 
The shared encoder $f_\phi$ maps the current observation to a latent $z_t = f_\phi(o_t)$, from which two parallel heads read out the modalities of interest: an action head $\pi_\theta(a\mid z)$ producing the chunk $a_{t:t+k}\!\in\!\mathbb{R}^{k\times d_a}$, and a world-model head $g_\psi(s\mid z)$ producing a future state $s_{t+T}$ at horizon $T$, \textit{fused within the trunk}:
\begin{equation}
p_{\theta,\psi}\!\big(a_{t:t+k},\, s_{t+T} \,\big|\, o_t\big)
= p_\psi\!\big(s_{t+T} \mid z_t\big)\,
  p_\theta\!\big(a_{t:t+k} \mid z_t\big),
\qquad z_t = f_\phi(o_t),
\end{equation}
with both heads trained jointly under
$\mathcal{L}_{\text{WAM}} = \mathcal{L}_{\text{action}}(a_{t:t+k}) + \lambda\, \mathcal{L}_{\text{world}}(s_{t+T})$,
where $\lambda$ trades off action fidelity against future prediction supervision.
The key property is that $\mathcal{L}_{\text{world}}$ supervises the \emph{shared trunk} through dynamics produced by \emph{both} embodiments, while $\mathcal{L}_{\text{action}}$ supervises through aligned actions we exhausted in Sec.~\ref{subsec:aligned-action}. Human data can therefore shape the shared representation $z_t$ through future-scene prediction even when its action labels do not transfer faithfully. 

This reframing exposes the central question that prior WAM work has not systematically addressed: \emph{\textbf{what should $s_{t+T}$ be?}} The world target decides whether human and robot data converge to a shared representation or are pulled apart by embodiment-specific signal. We turn to this question next.

\vspace{-5pt}
\section{\egowam: A Controlled Study of World Representations}
\label{sec:egowam}
\vspace{-5pt}

For WAM co-training to transfer from egocentric human data to robot manipulation, we posit three desiderata for the world representation:
\begin{itemize}[leftmargin=*,topsep=0pt,itemsep=0pt,parsep=0pt]
\item \textbf{(D1) Appearance abstraction.} Targets rewarding photometric reconstruction force the trunk to encode embodiment-specific appearance, crowding out the structure that governs task outcomes.
\item \textbf{(D2) Cross-embodiment consistency.} Targets should represent the \emph{effect} rather than the \emph{agent}, so a human hand and a robot gripper producing similar physical change induce similar supervision.
\item \textbf{(D3) Ego-motion factoring.} Image-coordinate targets conflate head rotation with scene change, giving the same event different supervision under moving human versus static robot cameras.
\end{itemize}

These desiderata define the axis \egowam studies; we instantiate three targets along it---pixel VAE, DINO features, and 3D flow. The overview of our model architecture is shown in Fig.~\ref{fig:method}. Further implementation details are provided in Appendix~\ref{supp:sec:implement-details}.
\vspace{-5pt}

\subsection{Single-Backbone Architecture and Controlled Variables}
\label{subsec:architecture}

\egowam builds on the Heterogeneous Pretrained Transformer (HPT) backbone~\cite{wang2024scalingproprioceptivevisuallearningheterogeneous, punamiya2026egoverseegocentrichumandataset}, where embodiment-specific stems tokenize each input into a shared latent space and a single transformer trunk operates on the unified token stream. This inductive bias extends naturally to WAMs: observation, action, and future-prediction tokens coexist in one stream and attend within the same trunk, exposing future dynamics and action prediction as two read-outs of a shared latent. We inherit the tokenizer and trunk and add a second bank of learnable \emph{future tokens} and a \emph{swappable} world-model head that consumes them (Fig.~\ref{fig:method}).

\noindent \textbf{Heterogeneous Tokenizers.}
Each embodiment is tokenized through a shallow, embodiment-specific stem. A shared ego-vision stem encodes $I^{\text{ego}}_t$ with a ResNet-18 encoder~\cite{he2016resnet} or pretrained DINO encoder~\cite{oquab2023dinov2, simeoni2025dinov3}; a matching wrist-vision stem encodes $I^{\text{wrist}}_t$ on robot batches. The proprioception stem encodes the end-effector pose with a per-embodiment MLP. Shared learnable \textit{action} and \textit{future} tokens then query the trunk, which routes them to the action and world-model heads.

\noindent \textbf{Action Head.}
The action head $\pi_\theta$ is a multi-block transformer decoder trained with conditional flow matching~\cite{lipman2023flowmatching, punamiya2026egoverseegocentrichumandataset}. Given a clean chunk $a_{t:t+k}$ and noise $\epsilon\sim\mathcal{N}(0,I)$, we draw $\tau\sim\text{Beta}(1.5,1.0)$ and form $a^\tau_{t:t+k}=(1-\tau)\epsilon + \tau\, a_{t:t+k}$ to initialize the action tokens, with $\tau$ embedded along the hidden dimension. Alternating self- and cross-attention blocks denoise the tokens while injecting trunk context, and a linear layer projects them into the action dimension. 

\noindent \textbf{Swappable World-Model Head.} 
The world-model head $g_\psi$ is conditioned on the trunk embeddings and predicts the target $s_{t+T}$ at the embodiment-specific horizon $T$ (Sec.~\ref{subsec:aligned-action}). 
We defer the choices of world-model head and target to Sec.~\ref{subsec:world-variants}, where we vary the world representation while holding the trunk, action head, and data mixture fixed.

\subsection{World-Model Target Instantiations}
\label{subsec:world-variants}

We instantiate three world targets spanning desiderata (D1)--(D3), each pairing a target $s$ with a suitable head and trained under a shared linear path $s^\tau = (1-\tau)\epsilon + \tau s$, $\tau\in[0,1]$, $\epsilon\sim\mathcal{N}(0,I)$.

\noindent \textbf{Pixel VAE (Reconstruction Baseline).}
The Pixel-VAE variant predicts a future ego frame in the latent space of a pretrained video VAE~\cite{wan2025} optimized for photometric reconstruction: $s = \mathrm{VAE}(I^{\text{ego}}_{t+T})$.
The head is a diffusion transformer (DiT) without text conditioning, either the pretrained VACE-1.3B~\cite{jiang2025vace} transformer (\textbf{Pixel-PT}) or a lightweight $6$-block DiT trained from scratch (\textbf{Pixel}; App.~\ref{supp:subsec:wm-head}), trained to predict noise: $\mathcal{L}^{\text{VAE}}_{\text{world}} = \mathbb{E}\big\|\epsilon - \epsilon_\psi(s^\tau, \tau, f_\phi(o))\big\|^2$.
This target violates all three desiderata and serves as our reconstruction-level baseline.

\noindent \textbf{DINO Features (Semantic Abstraction).} The target is DINO~\cite{oquab2023dinov2} patch features of the ego frame at $t+T$: $s = \mathrm{DINO}(I^{\text{ego}}_{t+T})$, replacing photometric fidelity with prediction in a semantically structured space. The head follows the wide-DDT design of RAE~\cite{zheng2025diffusiontransformersrepresentationautoencoders}, which widens the denoiser to match the high channel count of semantic latents and decouples noise-prediction from feature-conditioning blocks. The objective remains noise prediction: $\mathcal{L}^{\text{RAE}}_{\text{world}} = \mathbb{E}\big\|\epsilon - \epsilon_\psi(s^\tau, \tau, f_\phi(o))\big\|^2$. DINO features remain spatially indexed in image coordinates, so head ego-motion (D3) is only partially mitigated.

\noindent \textbf{3D Flow (Spatial Abstraction).}
The target is a dense 3D motion field over $[t, t+T]$, expressed in the camera-stabilized frame at time $t$: $s=F_{[t, t+T]}$. Raw 3D point displacement on egocentric video is dominated by ego-motion: stationary objects induce large apparent flow whenever the wearer turns their head. We address this by feeding the pretrained 3D point tracker~\cite{xiao2025spatialtrackerv23dpointtracking, lu2026track4world} with Aria VIO camera poses~\cite{engel2023projectarianewtool}, so the returned point positions $X_t, X_{t+T}$ share a consistent world frame. We then map the future position back to the camera frame at $t$: $\tilde X_{t+T} = (T^{\text{cam}}_t)^{-1}\, T^{\text{cam}}_{t+T}\, X_{t+T}$, and define the flow target as $s = F_{[t, t+T]} = \tilde X_{t+T} - X_t$. After stabilization, static background yields near-zero flow while manipulated objects retain motion proportional to physical displacement, abstracting dynamics from both appearance and viewpoint. 
The head is an alternating self/cross-attention decoder that takes query points $q$ uniformly sampled from the current ego frame as additional conditioning, and regresses velocity at those points: $\mathcal{L}^{\text{Flow}}_{\text{world}} = \mathbb{E}\big\|u_\psi(s^\tau, \tau, f_\phi(o), q) - (s_q - \epsilon_q)\big\|^2$.
This target satisfies all three desiderata by construction and is the most spatially grounded of the three.

\subsection{Joint Training and Action-Only Inference}
\label{subsec:training-inference}

\noindent \textbf{Joint Training.} \egowam is trained end-to-end under
\begin{equation}
\mathcal{L}_{\textsc{EgoWAM}} = \underbrace{\mathcal{L}^{\text{robot}}_{\text{action}} + \mathcal{L}^{\text{human}}_{\text{action}}}_{\mathcal{L}_{\text{action}}} + \lambda \underbrace{\big(\mathcal{L}^{\text{robot}}_{\text{world}} + \mathcal{L}^{\text{human}}_{\text{world}}\big)}_{\mathcal{L}_{\text{world}}},
\label{eq:egowam-loss}
\end{equation}
with $\lambda = 1$. Each step draws a mini-batch from $\mathcal{D}_R$ and $\mathcal{D}_H$; both pass through the shared tokenizers and trunk, after which the action head yields $\mathcal{L}_{\text{action}}$ (Eq.~\ref{eq:bc-cotrain}) and the world-model head yields $\mathcal{L}_{\text{world}}$ (Sec.~\ref{subsec:world-variants}). 
Both losses supervise the shared trunk $f_\phi$: when human action labels transfer weakly, world prediction compensates and shapes the trunk representation.

\noindent \textbf{Action-Only Inference.}
Recent work~\cite{yuan2026fastwamworldactionmodels, li2025unifiedvideoactionmodel} shows that WAMs benefit from \emph{training-time representation shaping} rather than test-time imagination. We therefore drop the world-model head at inference and unroll only the action head to decode $a_{t:t+k}$ from the trunk embeddings, matching the latency of a same-size BC policy (30~Hz) while retaining the cross-embodiment representation learned under joint supervision. This decoupling is central to our positioning: \egowam is an \emph{instrument for studying which world representation transfers}, with findings deployable at BC cost.

\begin{figure*}[h]
    \centering
    \includegraphics[width=\linewidth]{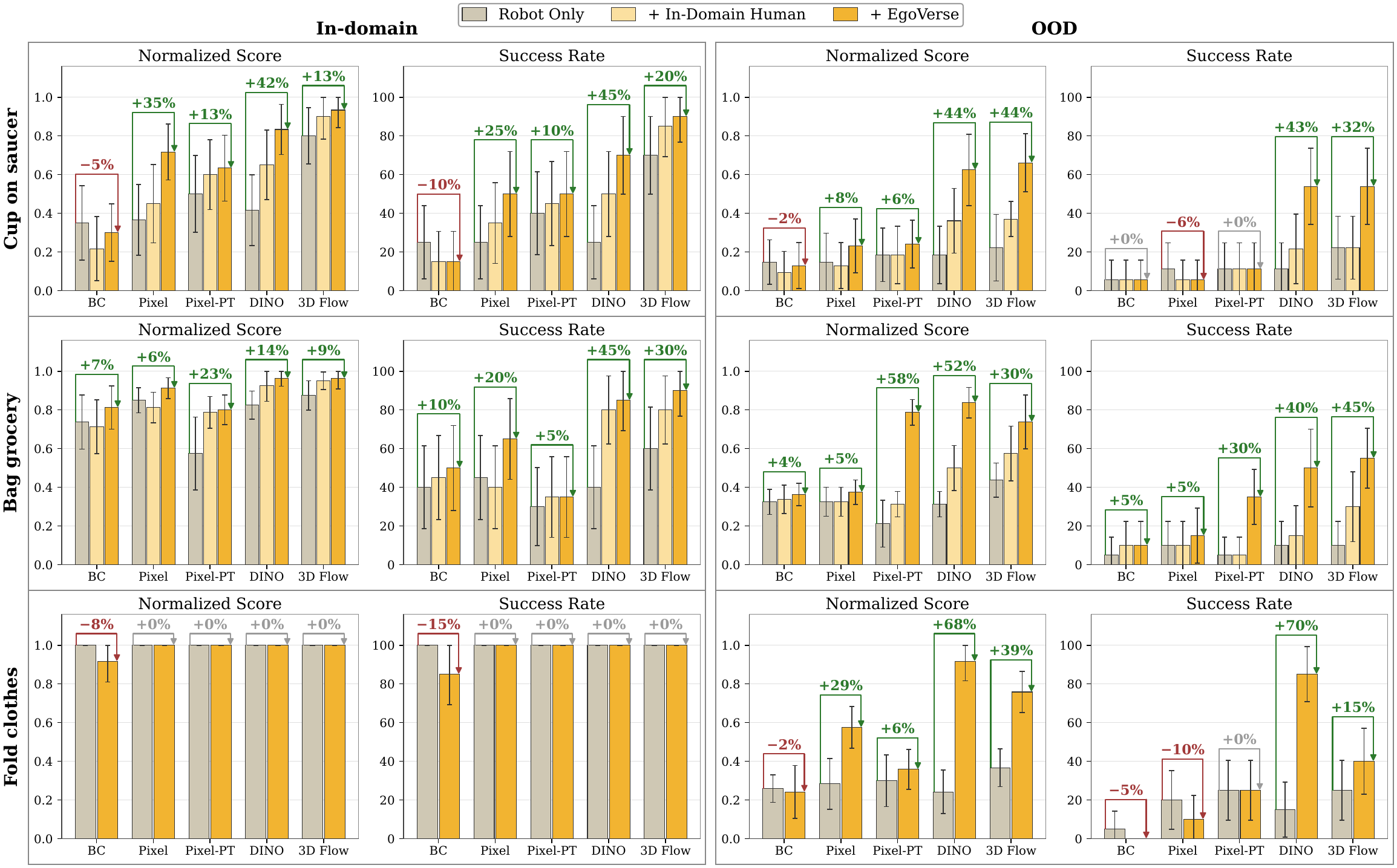}
    \vspace{-18pt}
    \caption{\small
    \textbf{Quantitative Comparison on Real-World Rollouts.} Normalized score and success rate across three bimanual tasks under ID and OOD evaluation, comparing BC against four WAM variants. WAM co-training consistently outperforms BC: human data often degrades BC yet yields large WAM gains. Pixel transfers weakly, while DINO drives the strongest OOD generalization and 3D Flow the best ID performance. 
    Error bars indicate 95\% confidence intervals ($\pm 1.96$ standard errors).
    }
    \label{fig:exp-main}
    \vspace{-15pt}
\end{figure*}

\begin{wrapfigure}{r}{0.50\textwidth}
\vspace{-35pt}
\centering
\includegraphics[width=\linewidth]{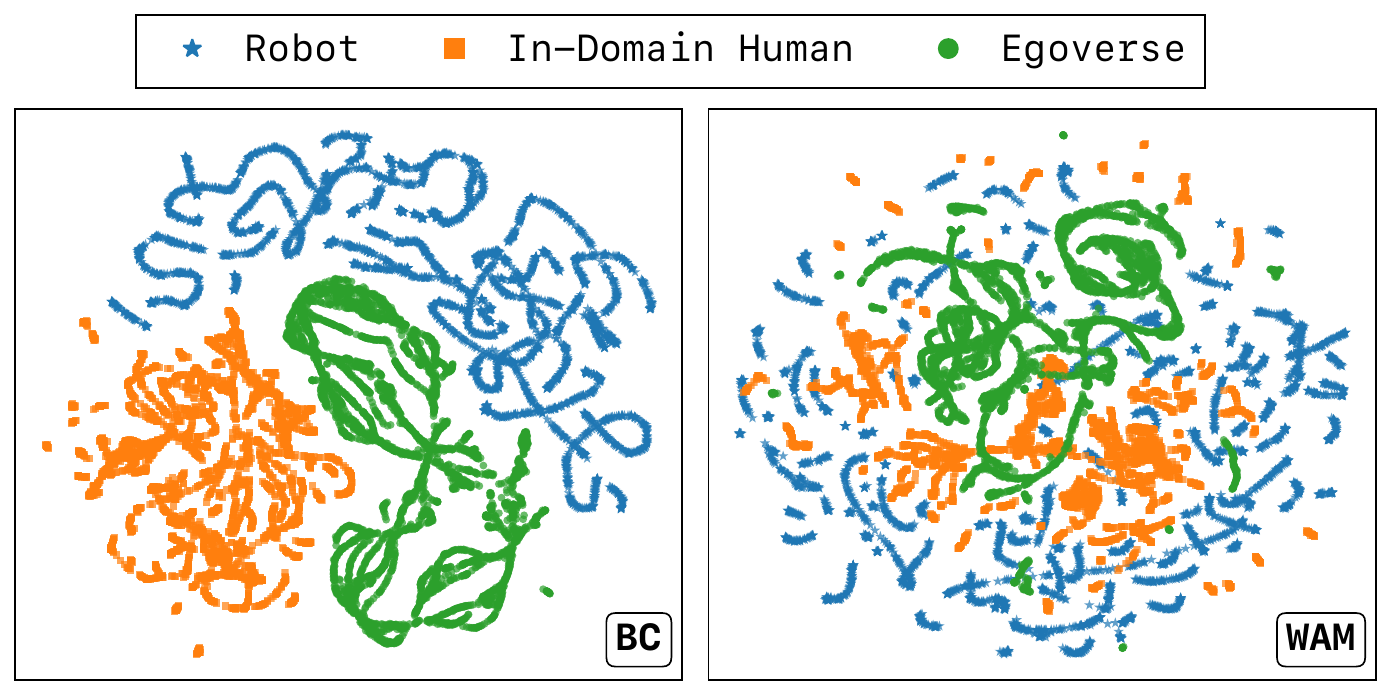}

\vspace{-8pt}
\caption{\small \textbf{UMAP of Trunk Embeddings on Cup-on-Saucer.} BC separates human-robot embeddings; WAM aligns them in a shared latent.}
\label{fig:exp-umap}
\vspace{-28pt}

\end{wrapfigure}
\begin{figure*}[t]
    \centering
    \includegraphics[width=\linewidth]{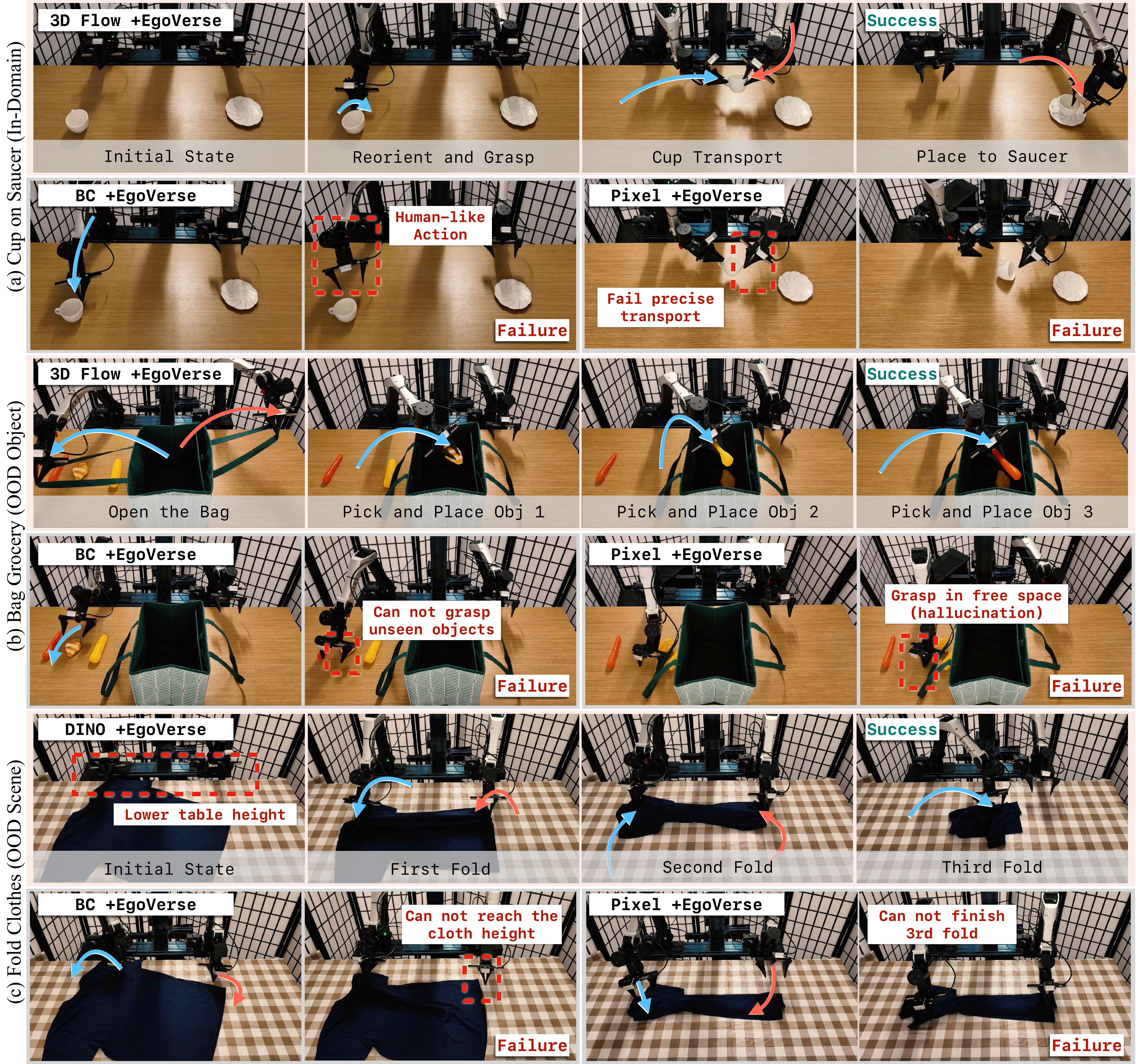}    
    \vspace{-18pt}
    \caption{\small
    \textbf{Qualitative Comparison on Real-World Rollouts.} WAM variants (3D Flow, DINO) are compared against BC and Pixel baselines under \textit{EgoVerse co-training}. \textbf{(a)} BC produces human-like motion, Pixel fails at precise cup transport; \textbf{(b)} BC cannot grasp unseen objects, Pixel hallucinates a free-space grasp; \textbf{(c)} BC overfits and fails to reach the cloth, Pixel's geometric confusion leaves the third fold incomplete.
    }
    \label{fig:exp-qual-wam}
    \vspace{-18pt}
\end{figure*}

\section{Experiments}
\label{sec:experiments}

We present a systematic study of \egowam for human-robot co-training, organized around three main questions: \textbf{Q1}: Does WAM co-training scale and transfer from large-scale in-the-wild human data better than BC co-training? \textbf{Q2}: Which world representation best enables transfer from human data to robot manipulation? \textbf{Q3}: How sensitive is each paradigm to human--robot alignment?

We provide additional experimental results and analysis in Appendix~\ref{supp:sec:exp-analysis} and robot-to-robot transfer simulation results in Appendix~\ref{supp:sec:sim}.

\subsection{Experimental Setup}
\label{subsec:exp-setup}

\noindent \textbf{Hardware Setup.}
Our bimanual robot platform has two upright-mounted 6-DoF ARX5 arms with parallel-jaw grippers, head-mounted Project Aria glasses~\cite{engel2023projectarianewtool} for egocentric RGB shared with human demonstrations, and two wrist-mounted Intel RealSense D405 cameras. Robot actions are per-arm 6-DoF end-effector poses with gripper state ($a^R_{t:t+k} \in \mathbb{R}^{k \times 14}$); human and robot data are unified into the camera-centered $\mathrm{SE}(3)$ action space and quantile-normalized (Sec.~\ref{subsec:aligned-action}).

\begin{wrapfigure}{r}{0.30\textwidth}
\vspace{-10pt}
\centering
\includegraphics[width=\linewidth]{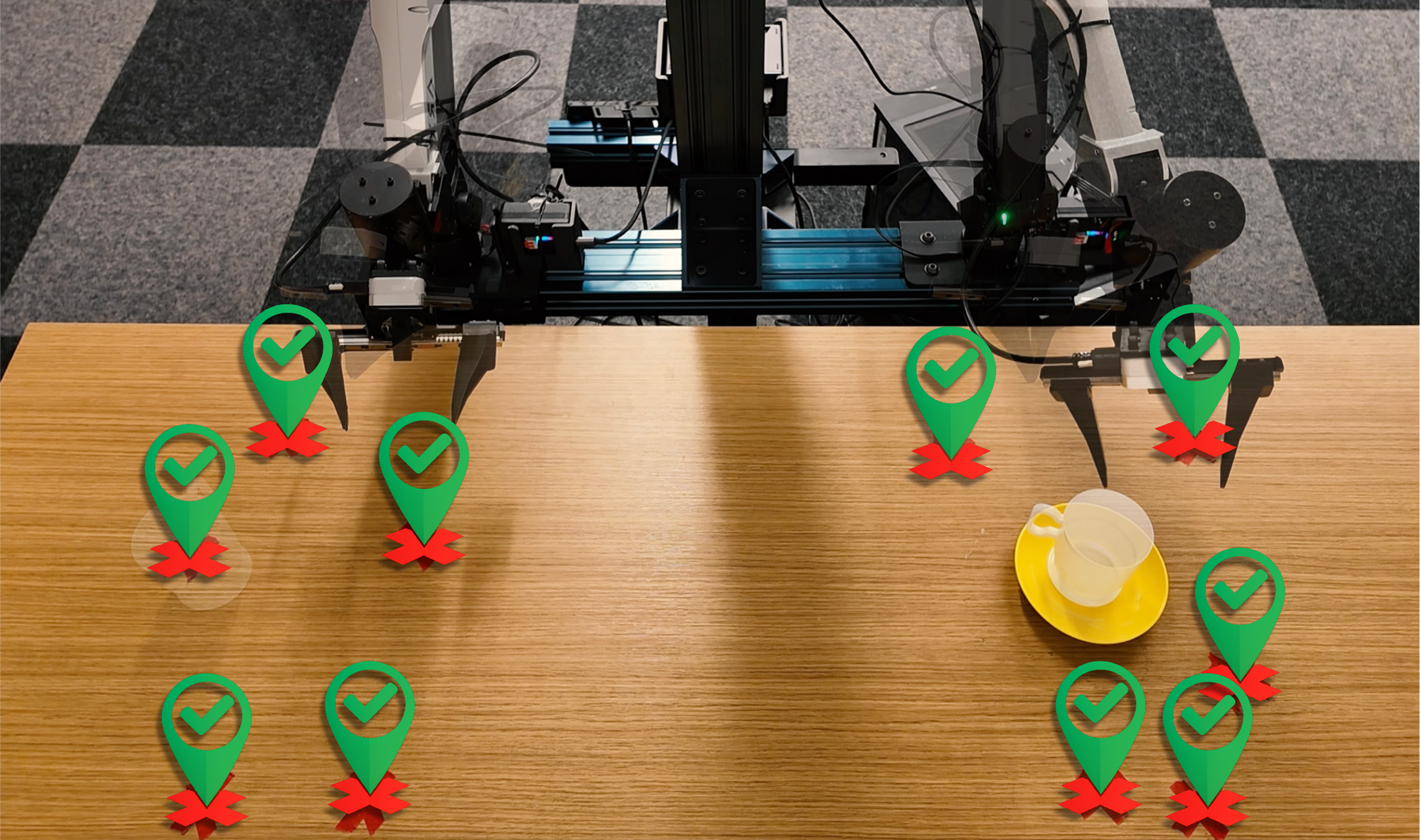}

\vspace{-8pt}
\caption{\small \textbf{Spatial Gains for 3D Flow.} 3D Flow succeeds (green) where BC fails (red) at cup positions across the workspace.}
\label{fig:exp-spatial}
\vspace{-15pt}

\end{wrapfigure}

\noindent \textbf{Tasks.}
We evaluate on three bimanual tasks from the EgoVerse flagship set~\cite{punamiya2026egoverseegocentrichumandataset}, spanning precise rigid-object manipulation, deformable manipulation, and long-horizon sequencing: 
(1) \textbf{cup-on-saucer}: Reorient a cup from a randomized pose and place it upright on a saucer at a randomized position;
(2) \textbf{fold-clothes}: Three-fold a T-shirt from random initial configurations;
(3) \textbf{bag-grocery}: Open a grocery bag and load three items into it from randomized positions.

\noindent \textbf{Data.}
We collect 300--360 robot demonstrations per task via Meta Quest 3, with randomized placements, orientations, and 4--8 object combinations. Human data spans two regimes varying in \emph{scale} and \emph{observation alignment}: (1) \textbf{In-Domain Human} (\textbf{1:1} with robot data): same scenes and objects as the robot, but unmatched viewpoint and behavior; (2) \textbf{EgoVerse} (\textbf{${\sim}$10:1}): the full EgoVerse-A flagship split per task~\cite{punamiya2026egoverseegocentrichumandataset}, with diverse scenes, objects, and demonstrators and no deliberate scene and object alignment.

\noindent \textbf{Evaluation Protocol.}
Each method is evaluated on \textbf{ID} (20 rollouts: seen objects and scene, randomized positions and orientations) and \textbf{OOD} (20 rollouts: 10 unseen objects in the training scene, 10 seen objects in novel scenes with varied backgrounds and table heights). We report normalized sub-task score and success rate over \textbf{1800} total real-world rollouts.

Further details on the experimental setup are provided in Appendix~\ref{supp:sec:real-world-exp-details}.

\subsection{Core Results and Findings}
\label{subsec:exp-core-findings}

\noindent \textbf{(Q1) WAM vs. BC Co-Training under Natural Human Data.}
When the robot and human action patterns are not well aligned---for example, the human hand tends to hold the cup sideways while the gripper grasps it from the front (Fig.~\ref{fig:teaser})---BC degrades by reproducing these \textit{inexecutable human-like actions} (Fig.~\ref{fig:exp-qual-wam}), falling below its robot-only baseline on cup-on-saucer and fold-clothes (Fig.~\ref{fig:exp-main}), whereas the DINO and 3D-flow WAMs consistently turn the same data into reliable gains. Both also improve further from In-Domain Human to the larger and more diverse EgoVerse data, most clearly on OOD evaluation.
Moreover, even when large amounts of human data are introduced, BC tends to \textit{overfit} to the robot data alone and fails to benefit from human data for generalization: on fold-clothes, BC$+$EgoVerse cannot adapt to novel scenes with lower table heights, whereas the DINO-based WAM leverages in-the-wild human data to achieve better generalization to new objects and scenes (Fig.~\ref{fig:exp-qual-wam}). Fig.~\ref{fig:exp-umap} traces this to representation: BC isolates human and robot (and even human--human) embeddings under unaligned actions, whereas WAM aligns them into a shared space through additional task-relevant dynamics supervision. An ablation of the human supervision on cup-on-saucer further shows that the two channels are complementary: 3D-flow supervision alone on human data outperforms action supervision alone, and combining both performs best (App.~\ref{supp:subsec:ablation-human-modality}). 

\begin{wrapfigure}{l}{0.60\textwidth}
\vspace{-10pt}
\centering
\includegraphics[width=\linewidth]{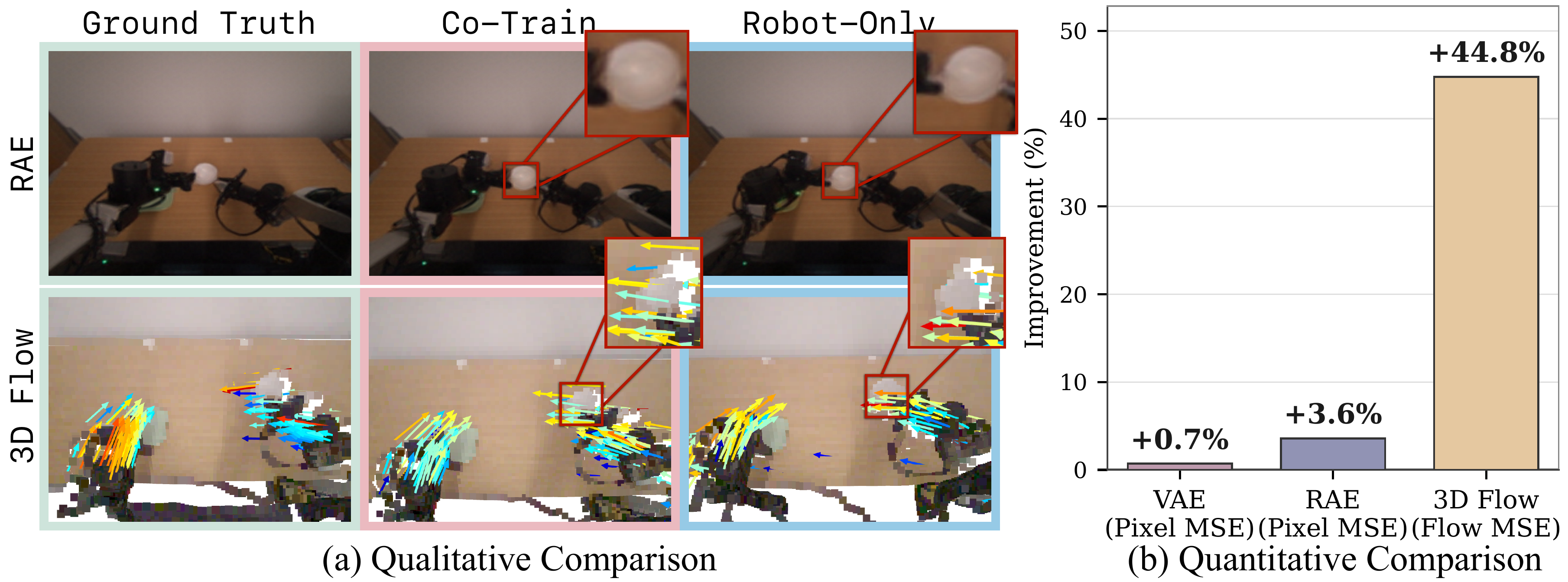}

\vspace{-5pt}
\caption{\small \textbf{World Prediction Comparison.} \textbf{(a)} Co-training improves RAE and 3D Flow predictions, with better \textit{object shape and motion}; \textbf{(b)} 3D Flow's gains far exceed Pixel and DINO.}
\label{fig:exp-world-pred}
\vspace{-10pt}

\end{wrapfigure}
\noindent \textbf{(Q2) World Representation Choice for Cross-Embodiment Transfer.}
The world representation is the next critical axis. Fig.~\ref{fig:exp-main} shows that pixel-based prediction transfers weakly: reconstructing raw appearance entangles embodiment- and scene-specific detail that does not transfer, leaving the hallucination and geometric confusion in Fig.~\ref{fig:exp-qual-wam}: a free-space grasp on bag-grocery and an unfinished final fold on fold-clothes. A pretrained video prior (Pixel-PT) does not fix this: with robot data alone, it even hallucinates a bag opening before the robot acts (App.~\ref{supp:subsec:failure-pt}), and neither its far larger head (App.~\ref{supp:subsec:head-capacity}) nor denser multi-frame prediction (App.~\ref{supp:subsec:pixel-pt-dense}) closes the gap, suggesting it lies in the world representation itself.
Abstraction fixes this in two complementary ways: the DINO-based WAM predicts semantic features and yields the strongest average OOD generalization to unseen objects and scenes (Fig.~\ref{fig:exp-main}), whereas the 3D-flow WAM predicts camera-stabilized motion and delivers the best in-domain results, succeeding at precise cup placement across the workspace (Fig.~\ref{fig:exp-spatial}). 
Fig.~\ref{fig:exp-world-pred} grounds this contrast: pixel (VAE) prediction barely changes, the semantic (RAE) target refines \textit{object shape} from human data but only moderately, and the 3D-flow target benefits most, recovering \textit{object motion} that robot-only prediction leaves static.

\begin{wrapfigure}{l}{0.65\textwidth}
\centering
\includegraphics[width=\linewidth]{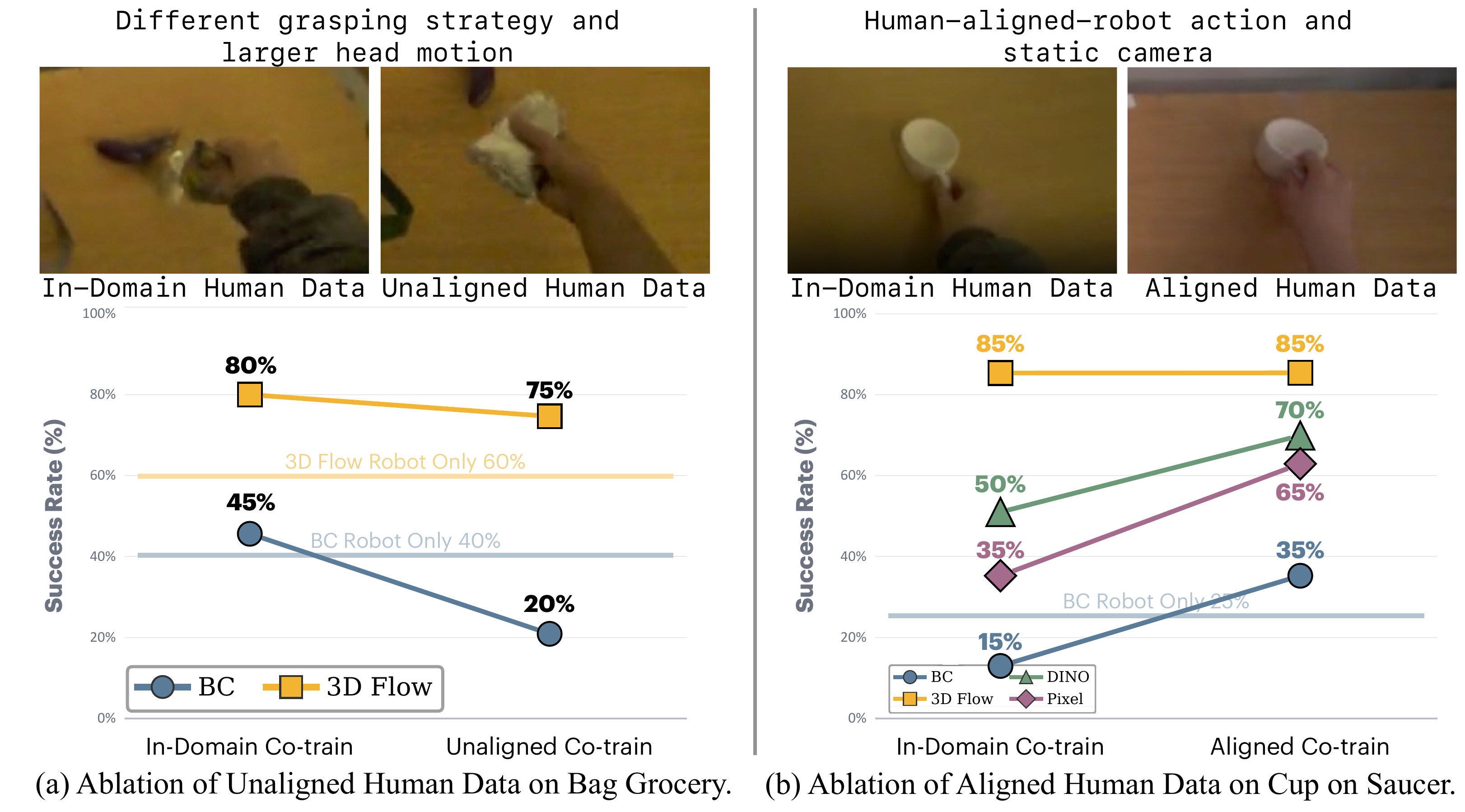}

\vspace{-8pt}
\caption{\small \textbf{Ablation of Human--Robot Alignment.} \textbf{(a)} Unaligned human data collapses BC below robot-only; 3D flow stays robust. \textbf{(b)} Aligned human data lifts BC above robot-only and improves Pixel/DINO; 3D flow holds either way. In-domain success rate; horizontal lines: robot-only baselines.}
\label{fig:exp-ablation-aligned-human}
\vspace{-10pt}

\end{wrapfigure}

\noindent \textbf{(Q3) Ablation: The Role of Human--Robot Alignment.}
We further study the hypothesis that unaligned human data degrades BC (Fig.~\ref{fig:exp-ablation-aligned-human}; collection details in App.~\ref{supp:subsec:human-data}).

\noindent\hangindent=1em\hangafter=1\makebox[1em][l]{\textbullet}\textit{(a) Unaligned human data.} Among the three tasks, only bag-grocery benefits from action-level co-training (Fig.~\ref{fig:exp-main}), because its pick-and-place motions are naturally aligned between human and robot. As a counterfactual, we collect unaligned human demonstrations in which the human picks objects in an unusual manner, with a different grasping strategy and larger head motion, that maps to inexecutable robot actions. Under this data, BC collapses below the robot-only baseline, whereas the 3D-flow WAM stays robust and still surpasses robot-only.

{\parfillskip=0pt \noindent\hangindent=1em\hangafter=1\makebox[1em][l]{\textbullet}\textit{(b) Aligned human data.} We then probe the other extreme on cup-on-saucer, the task with the strongest human--robot mismatch (Fig.~\ref{fig:teaser}), where the demonstrator mirrors the robot's viewpoint\par}

\noindent and grasp strategy. Aligned human data lifts BC above its robot-only baseline, confirming that BC benefits from human data \emph{only} when the action distribution matches the robot's, naturally (bag-grocery) or by manual curation. Pixel and DINO also improve once human head motion is removed at collection time, showing how much image-coordinate targets suffer when D3 is violated. In contrast, the 3D-flow WAM is invariant to alignment: its camera-stabilized target factors out ego-motion by construction, so the gains others recover through manual alignment come for free.

\vspace{-5pt}
\section{Conclusion}
\label{sec:conclusion}
\vspace{-5pt}

We introduce \egowam, a human-robot WAM co-training framework studying what \textit{world representation} enables cross-embodiment transfer. It scales with \textit{in-the-wild human data} where action-only BC stalls, and the world representation proves critical: pixels transfer weakly, DINO drives object and scene generalization, and 3D flow grounds in-domain spatial gains. This yields a recipe for further study: a target that abstracts appearance, keeps effects consistent across embodiments, and factors out ego-motion, positioning human data as a flywheel for scalable robot learning.
We offer this study not as a definitive answer but as a point of departure, turning ``use human data'' into a concrete design axis for scaling robot manipulation toward open-world generalization.

\section{Limitations}
\label{sec:limitations}

Three limitations open future directions. 
\textbf{(1) Motion generalization.} Our gains stay at the context level; learning novel motion primitive / skill from human data (e.g., folding shorts from a T-shirt model) remains out of reach, and a more unified action representation is left to future study.
\textbf{(2) Multi-task scaling.} We isolate world representation with one policy per task; multi-task co-training on large-scale in-the-wild human data is promising to explore.
\textbf{(3) Open world representation.} 
Our study is a starting point, showing DINO and 3D-flow beat pixels for cross-embodiment transfer; the ``best'' world representation in scaling robot learning remains a valuable open research question.

\acknowledgments{%
This work was supported in part by the Toyota Research Institute through the TRI University 3.0 program.
We thank the members of the Robot Learning and Reasoning Lab (RL2) for helpful discussions and hardware support. 
}

\bibliography{references}

\clearpage
\appendix
\input{supplementary/supplementary}

\end{document}